\documentclass{article}

\usepackage{microtype}
\usepackage{graphicx}
\usepackage{subfigure}
\usepackage{booktabs} 

\usepackage{soul}
\usepackage{hyperref}
\usepackage{amsfonts} 
\usepackage{url}
\usepackage{floatflt}
\usepackage{graphicx} 
\usepackage[T1]{fontenc}    
\usepackage{tikz}
\usepackage{ctable}
\usepackage{multicol}

\usepackage{multirow}
\usepackage{subcaption}
\usepackage{wrapfig}
\usepackage{hyperref}

\usepackage[accepted]{icml2023}

\usepackage{amsmath}
\usepackage{amssymb}
\usepackage{mathtools}
\usepackage{amsthm}

\usepackage[capitalize,noabbrev]{cleveref}

\usepackage{booktabs} 
\usepackage{colortbl}
\def\redc{\cellcolor[HTML]{FF999A}}
\def\orangec{\cellcolor[HTML]{FFCC99}}
\def\yellowc{\cellcolor[HTML]{FFF8AD}}

\theoremstyle{plain}

\theoremstyle{definition}

\theoremstyle{remark}

\usepackage[textsize=tiny]{todonotes}

\def\our{Viewing Direction Gaussian Splatting}
\def\ourshort{VDGS}

\def\G{\mathcal{G}}
\def\N{\mathcal{N}}
\def\m{\mathrm{m}}
\def\x{ {\bf x} }
\def\F{\mathcal{F}}

\def\F{\mathcal{F}}

\def\G{\mathcal{G}}
\def\N{\mathcal{N}}
\def\m{\mathrm{m}}
\def\x{ {\bf x} }
\def\F{\mathcal{F}}

\def\our{Viewing Direction Gaussian Splatting}
\def\ourshort{VDGS}

\def\G{\mathcal{G}}
\def\N{\mathcal{N}}
\def\m{\mathrm{m}}
\def\x{ {\bf x} }
\def\d{ {\bf d} }
\def\F{\mathcal{F}}

\icmltitlerunning{}

\begin{document}

\twocolumn[
\icmltitle{Gaussian Splatting with NeRF-based Color and Opacity }



\icmlsetsymbol{equal}{*}

\begin{icmlauthorlist}
\icmlauthor{Dawid Malarz}{equal,yyy}
\icmlauthor{Weronika Smolak}{equal,yyy}
\icmlauthor{Jacek Tabor}{yyy}
\icmlauthor{Sławomir Tadeja}{comp}
\icmlauthor{Przemysław Spurek}{yyy}
\end{icmlauthorlist}

\icmlaffiliation{yyy}{Faculty of Mathematics and Computer Science, Jagiellonian University, Krakow, Poland}
\icmlaffiliation{comp}{Department of Engineering,
University of Cambridge, Cambridge, UK}

\icmlcorrespondingauthor{Przemysław Spurek}{przemyslaw.spurek@uj.edu.pl}

\icmlkeywords{Machine Learning, ICML}

\vskip 0.3in
]



\printAffiliationsAndNotice{\icmlEqualContribution} 

\begin{abstract}
\textit{Neural Radiance Fields} (NeRFs) have demonstrated the remarkable potential of neural networks to capture the intricacies of 3D objects. By encoding the shape and color information within neural network weights, NeRFs excel at producing strikingly sharp novel views of 3D objects. Recently, numerous generalizations of NeRFs utilizing generative models have emerged, expanding its versatility. In contrast, \textit{Gaussian Splatting} (GS) offers a similar render quality with faster training and inference as it does not need neural networks to work. It encodes information about the 3D objects in the set of Gaussian distributions that can be rendered in 3D similarly to classical meshes. Unfortunately, GS are difficult to condition since they usually require circa hundred thousand Gaussian components. To mitigate the caveats of both models, we propose a hybrid model \textit{\our{}} (\ourshort{}) that uses GS representation of the 3D object's shape and NeRF-based encoding of color and opacity. Our model uses Gaussian distributions with trainable positions (i.e. means of Gaussian), shape (i.e. covariance of Gaussian), color and opacity, and a neural network that takes Gaussian parameters and viewing direction to produce changes in the said color and opacity. As a result, our model better describes shadows, light reflections, and the transparency of 3D objects without adding additional texture and light components.  \\
\url{https://github.com/gmum/ViewingDirectionGaussianSplatting}  
\end{abstract}

\begin{figure}[!t]
    \begin{center}
    
    \includegraphics[width=0.45\textwidth]{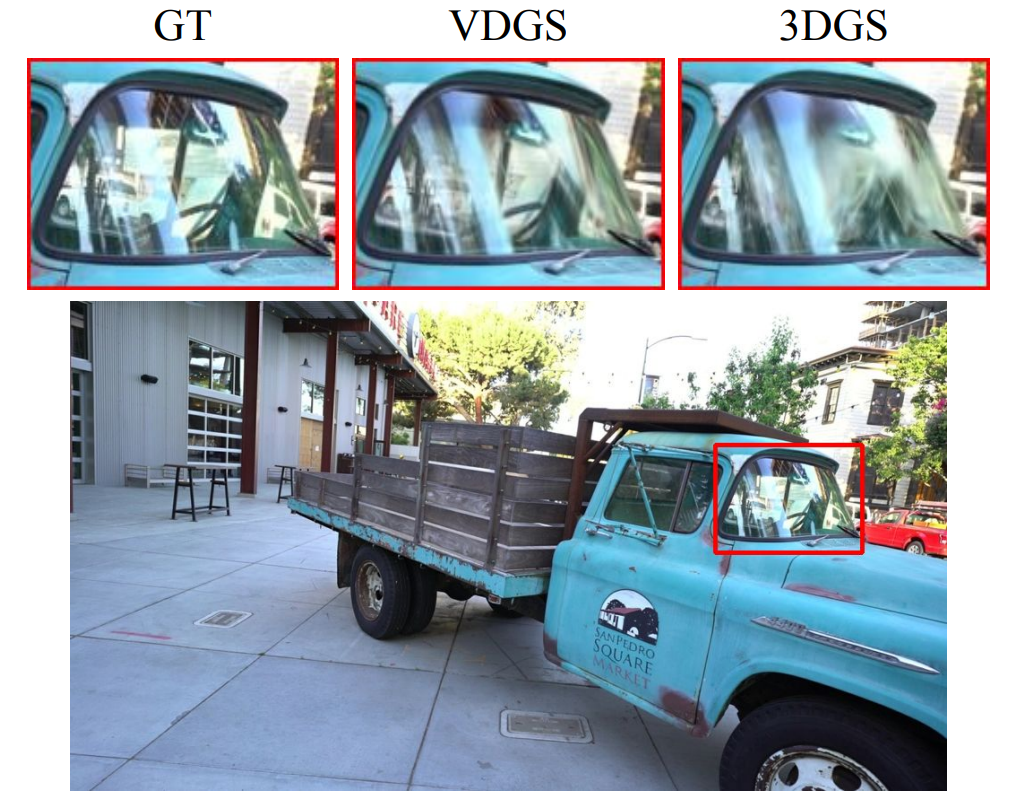} 
    \end{center}
    \caption{ Comparison of Gaussian Splatting (GS) and \our{} (VDGS). In our model, the color and opacity of Gaussians depend on the viewing direction. Consequently, we better model light reflection, and transparency of 3D objects (see elements in red rectangles).}
    \label{fig:merf_appro}
    \vspace{-0,4cm}
\end{figure}

\section{Introduction}

Over recent years, neural rendering has become a significant and, at the same time, prolific area of computer graphics research. The groundbreaking concept of \textit{Neural Radiance Fields} (NeRFs)~\cite{mildenhall2020nerf} has revolutionized 3D modeling, enabling the creation of complex, high-fidelity 3D scenes from previously unseen angles using only a minimal set of planar images and corresponding camera positions. This particular neural network architecture harnesses the connection between these images and fundamental methods and techniques of computer graphics (e.g. ray tracing) to conjure high-quality scenes from previously unseen vantage points.

\begin{figure*}
    \begin{center}
    \includegraphics[width=0.85\textwidth]{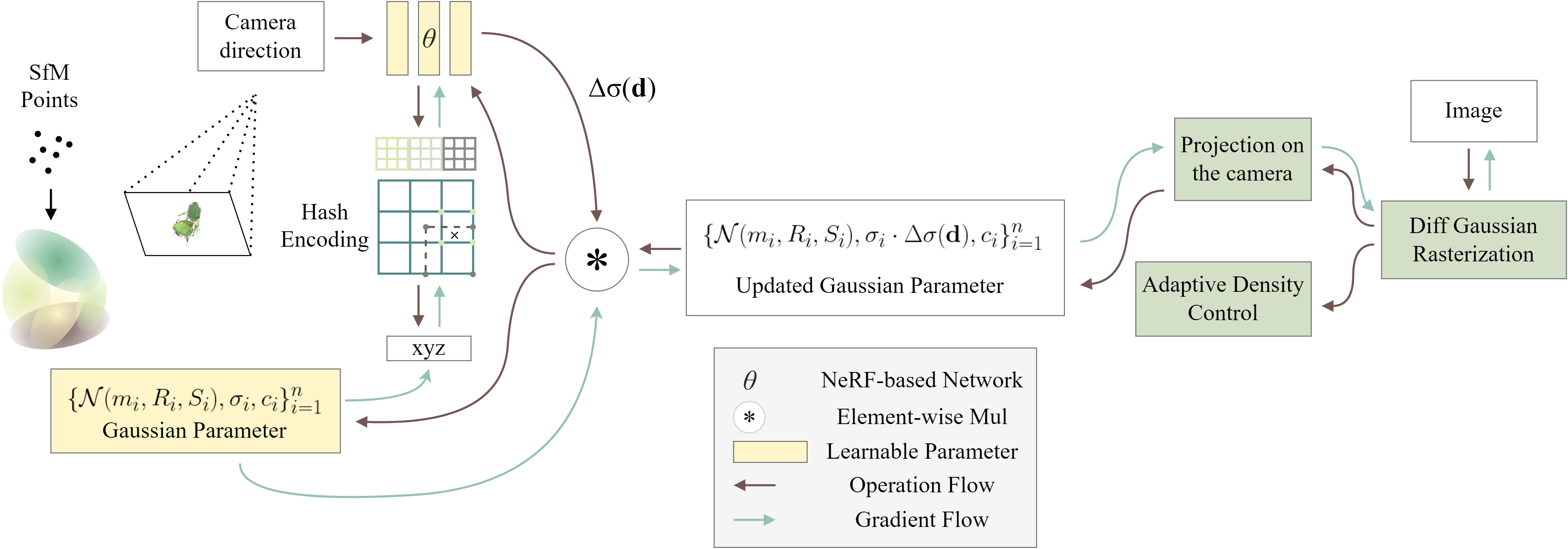} 
    \end{center}   
    \caption{ The optimization procedure starts with Structure from Motion (SfM) points, either sourced from COLMAP or created randomly, which establish the initial conditions for the 3D Gaussians. The camera position and Gaussian center, encoded with a hash, are fed into an MLP network to update the opacity $\delta \sigma$ of the 3D Gaussians in canonical space. Subsequently, a rapid differential Gaussian rasterization pipeline is employed to concurrently optimize both the MLP and the parameters of the 3D Gaussians. }
    \label{fig:schemma}
\end{figure*}

In NeRFs~\cite{mildenhall2020nerf}, the scene is represented with the help of fully connected network architectures. These networks take, as input, 5D coordinates consisting of camera positions and spatial locations. From this data, they are able to generate the color and volume density of each point within the scene. The loss function of NeRF draws inspiration from conventional volume rendering techniques~\cite{kajiya1984ray}, which involve rendering the color of every ray traversing the scene. In essence, NeRF embeds the shape and color information of the object within the neural network's weights.



NeRF architecture excels at generating crisp renders of new viewpoints in static scenes. However, it faces several limitations, primarily stemming from the time-consuming process of encoding object shapes into neural network weights. In practice, training and inference with NeRF models can take extensive time, making them often impractical for real-time applications.

\begin{figure}
    \centering
    \includegraphics[width=0.4\textwidth, clip]{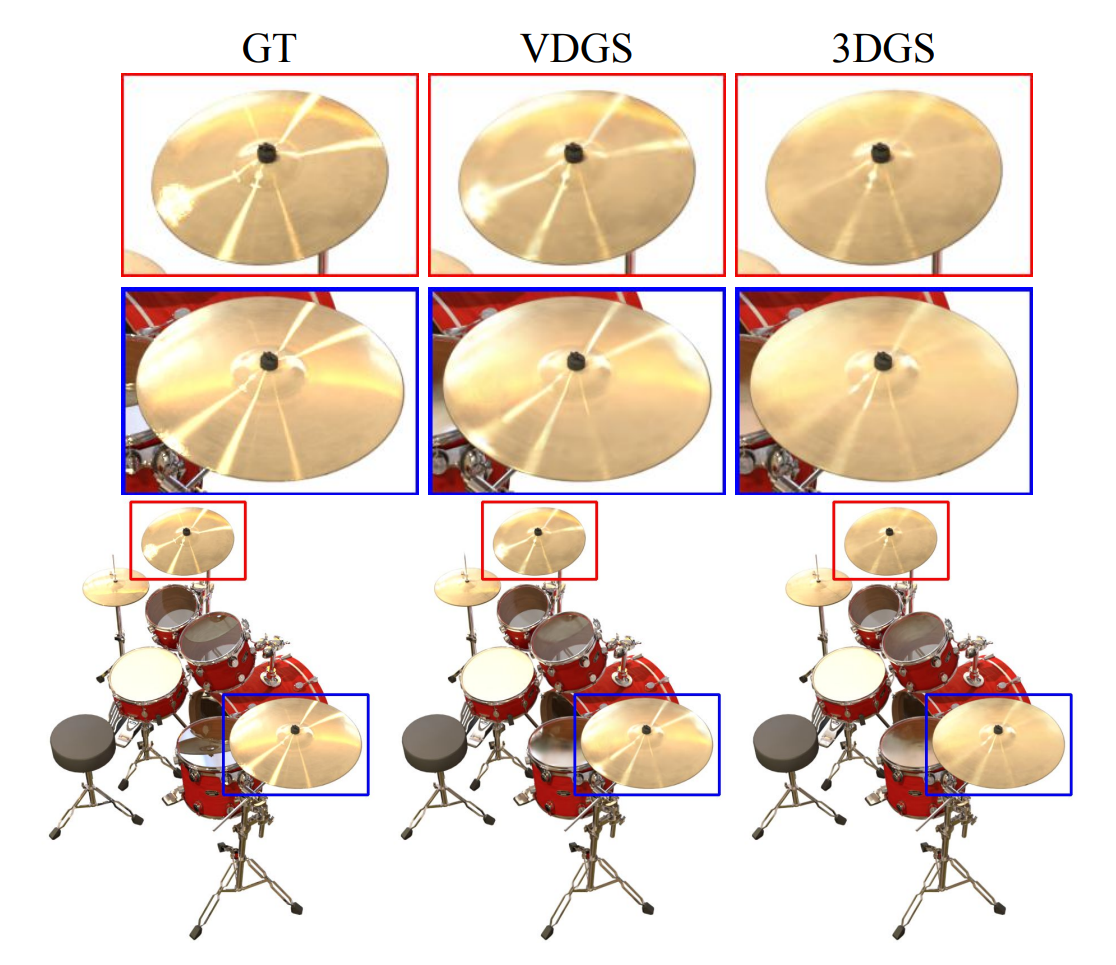}
    \caption{Visual comparison between classical GS and \our{} on the dataset:  Synthetic NeRF \cite{mildenhall2020nerf}. It further showcases superior performance in modeling shiny surfaces compared to the original GS.}
    \label{fig:shiny3}
\end{figure}

In comparison, Gaussian Splatting (GS) \cite{kerbl20233d} provides a similar quality of renders with more rapid training and inference. This is a consequence of GS not requiring neural networks. Instead, we encode information about the 3D objects in a set of Gaussian distributions. These Gaussians can then be used in a similar manner to classical meshes. Consequently, GS can be swiftly developed when needed to, for example, model dynamic scenes~\cite{wu20234d}. Unfortunately, GS is hard to condition as it necessitates a hundred thousand Gaussian components.

Both these rendering methods, i.e. NeRFs and GS, offer certain advantages and possess a range of caveats. In this paper, we present \textit{\our{}} (\ourshort{})--a new, hybrid approach that uses GS representation of the 3D object's shape and simultaneously employs NeRF-based encoding of color and opacity (see Figure~\ref{fig:merf_appro}). Our model uses Gaussian distributions with trainable positions (i.e., means of Gaussian), shape (i.e., the covariance of Gaussian), color (i.e., spherical harmonics), and opacity, as well as a neural network, which takes a position of Gaussian together with viewing direction to produce changes in opacity (see~Figure~\ref{fig:schemma}). \ourshort{} could also update to color or both color and opacity. However, during our ablation study, we found out that changing only opacity gave the most consistently high-quality results. Nonetheless, changes made to opacity indirectly also change color since other Gaussians can then be exposed and visible on the render.

\begin{figure*}
    \centering

    \includegraphics[width=0.8\textwidth, trim=0 0 0 0, clip]{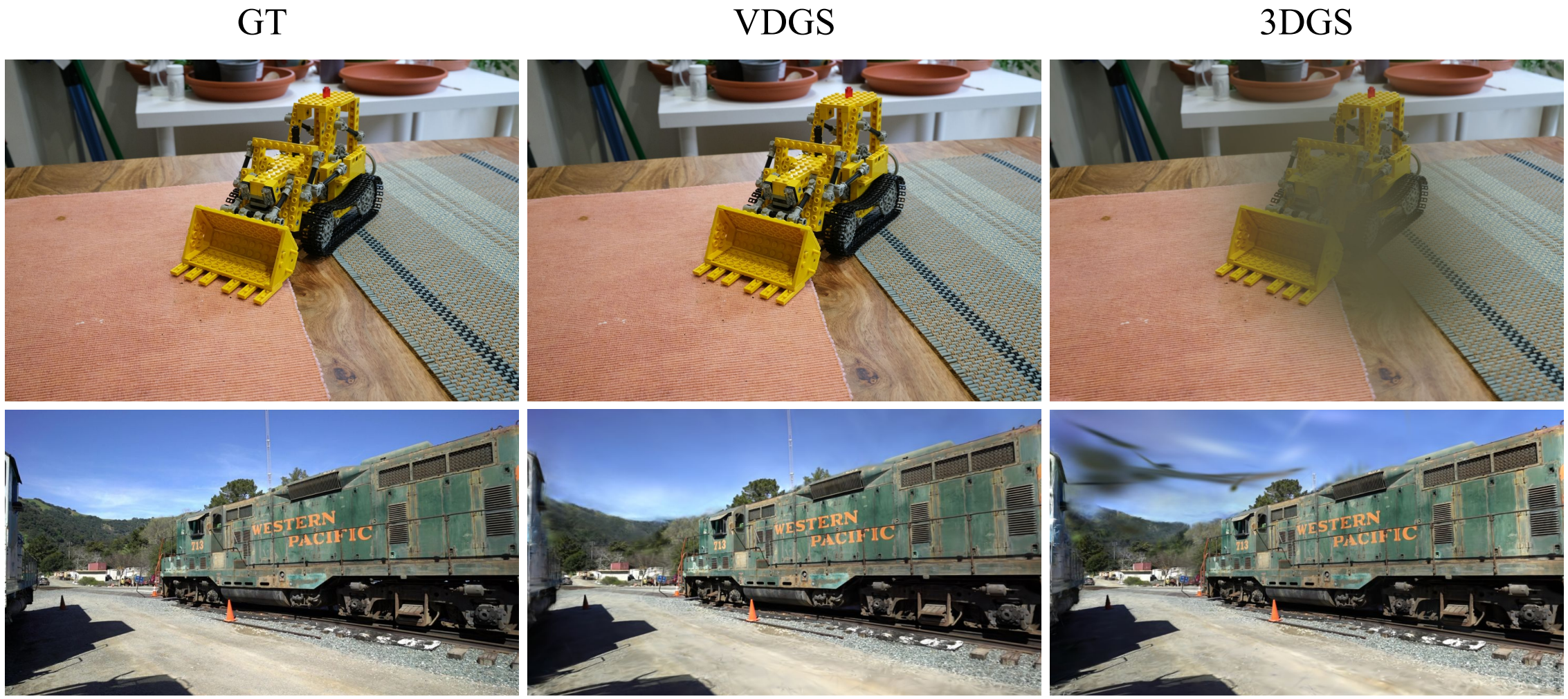}
    
    \caption{Visual comparison between classical GS and \ourshort{} on Tanks and Temples \cite{knapitsch2017tanks} and Mip-NeRF 360 \cite{barron2022mip} datasets. \ourshort{} compared to classical GS renders fewer artifacts which can be observed both on the renders as well as in the PSNR scores (see also Table \ref{tab:scene_mip_tt_db}).}
    \label{fig:artefacts_1}
\end{figure*}

\begin{figure}
    \centering
    \includegraphics[width=0.35\textwidth, clip]{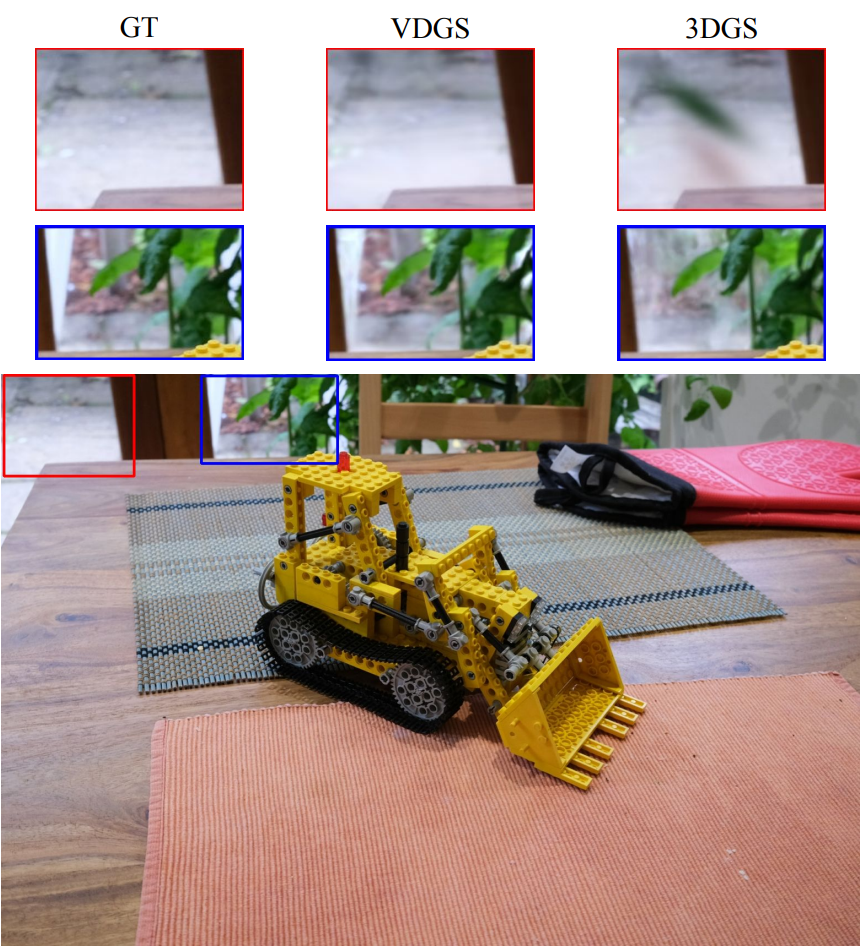}
    \caption{As visible in the render made using Mip-NeRF 360 scene \cite{barron2022mip}, \our{} not only performs well in reproducing an object in the main focus. It also better represents the glass surface and can eliminate artifacts.}
    \label{fig:shiny1}
\end{figure}

Such approaches inherit advantages of both NeRF and GS. First of all, \our{} has the same training and inference time as GS. Moreover, we can better model shadows, light reflection, and transparency in 3D objects by using viewing directions, as shown in Figure~\ref{fig:shiny3}. Additionally, due to the regulation of Gaussians' opacities, our model has the ability to reduce view-dependent artifacts, as shown in Figure \ref{fig:artefacts_1} and Figure \ref{fig:shiny1}. We do not use any physical model for texture or light. Therefore, our model is not a dedicated tool but a general solution that can be used in many general applications.

We can condition GS components effectively by controlling the colors and opacity of the Gaussian distributions. Such an approach could be used in many applications where NeRF is controlled by neural networks like  NeRF in the Wild, dynamic scenes, or generative models.



In summary, this work makes the following contributions:
\begin{itemize}
    \item We propose a hybrid architecture utilizing both NeRF and GS. We modulate the necessary changes to color and opacity using NeRF, whereas GS represents the 3D object's shape. Therefore, we inherit fast training and inference from GS and the ability to adapt to viewing direction from NeRF. 
    \item In contrast to GS, in the VDGS color and opacity are more sensitive to viewing direction without causing a notable increase in training time.   
    \item VDGS shows that GS can be conditioned by neural networks in a NeRF-based manner.
\end{itemize}

\section{Related Work}




Neural rendering is presently one of the most important areas of computer graphics. NeRF and its various generalizations \cite{barron2021mip,barron2022mip,liu2020neural,niemeyer2022regnerf,roessle2022dense,tancik2022block,verbin2022ref} use differentiable volumetric rendering to generate novel views of a static scene.
The prolonged training time associated with traditional NeRFs has been tackled before with the introduction of Plenoxels \cite{fridovich2022plenoxels}. In this approach, we replace the dense voxel representation with a sparse voxel grid, significantly reducing the computational demands. Within this sparse grid, density and spherical harmonics coefficients are stored at each node, enabling the estimation of color based on trilinear interpolation of neighboring voxel values. On the other hand, \citet{sun2022direct} delved into optimizing the features of voxel grids to achieve rapid radiance field reconstruction. \citet{muller2022instant} refined this approach by partitioning the 3D space into independent multilevel grids to enhance its computational efficiency further. \citet{chen2022tensorf} studies representation 3D object using orthogonal tensor components. A small MLP network extracts the final color and density from these components using orthogonal projection. Additionally, some methods use supplementary information like depth maps or point clouds \cite{azinovic2022neural, deng2022depth, roessle2022dense}. \citet{zimny2023multiplanenerf} proposed a MultiPlane version of NeRF with image-based planes to represent 3D objects. The TriPlane representation from EG3D~\cite{chan2022efficient} has a wide range of applications to diffusion-based generative model \cite{ssdNeRF}.


Alternatively, Gaussians have been widely used in a variety of fields, including shape reconstruction \cite{keselman2023flexible}, molecular structures modeling \cite{blinn1982generalization}, or the substitution of point-cloud data \cite{eckart2016accelerated}. Moreover, these representations can be employed in shadow \cite{nulkar2001splatting} and cloud rendering applications \cite{man2006generating}. 

Recently, 3D Gaussian Splatting (3D-GS) \cite{kerbl20233d}, which combines the concept of point-based rendering and splatting techniques for rendering, has achieved real-time speed with a rendering quality that is comparable to the best MLP-based renderer, namely, the Mip-NeRF \cite{barron2022mip}.

GS offers faster training and inference time than NeRF. Moreover, we do not use a neural network, as all the information is stored in 3D Gaussian components. Therefore, such a model finds application in dynamic scene modeling. Moreover, it is easy to combine with a dedicated 3D computer graphics engine \cite{kerbl20233d}. On the other hand, it is hard to condition GS since we typically have hundreds of thousands of Gaussian components. 

\begin{figure*}
    \centering

    \includegraphics[width=0.8\textwidth, trim=0 0 0 0, clip]{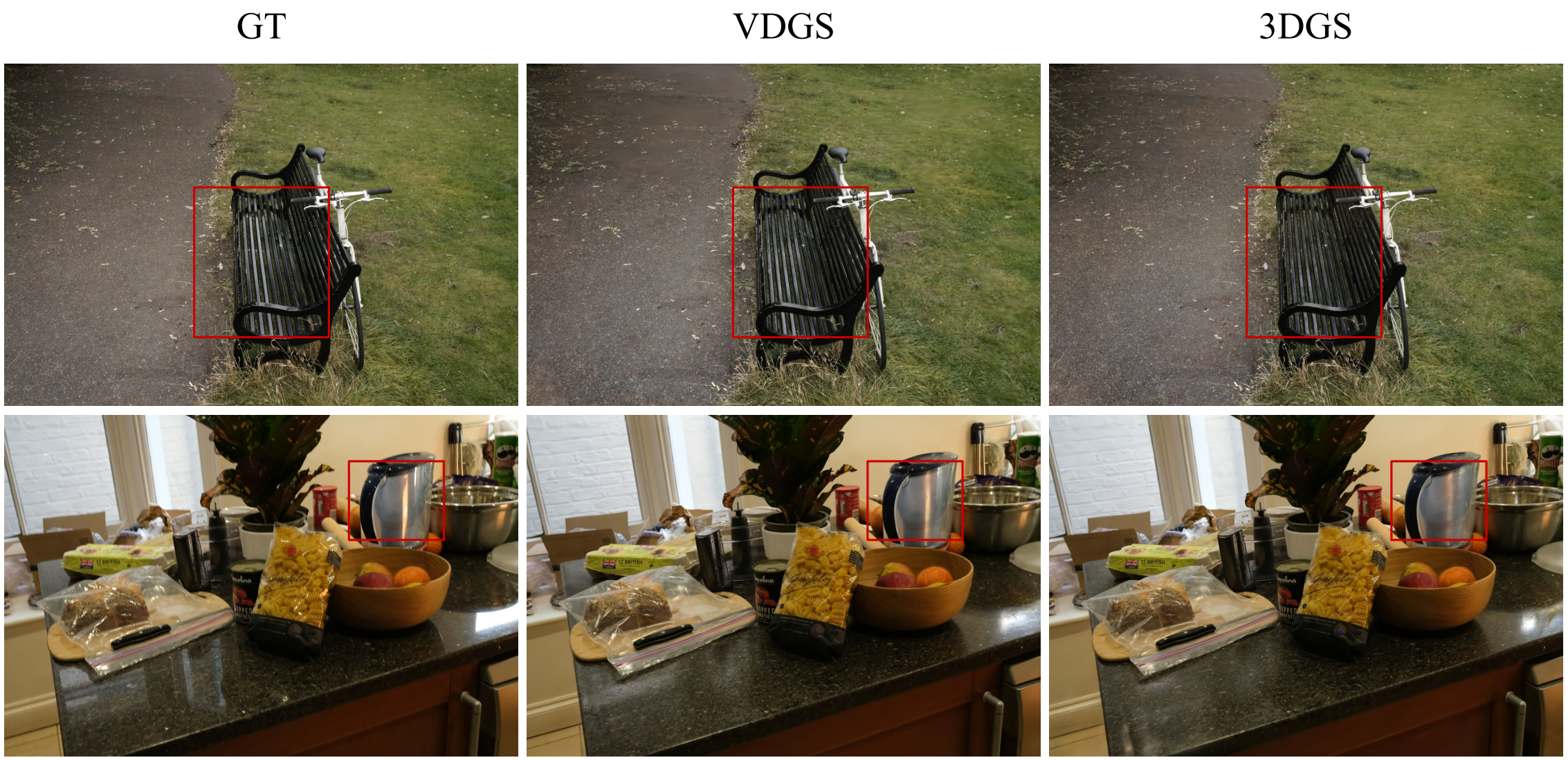}
    
    \caption{Example of a more accurate render of glass reflections made by \our{} compared to GS on the scene from Tanks and Temples \cite{knapitsch2017tanks}.}
    \label{fig:shiny_real}
\end{figure*}

\begin{figure}[b]
    \centering
    \includegraphics[width=0.4\textwidth, clip]{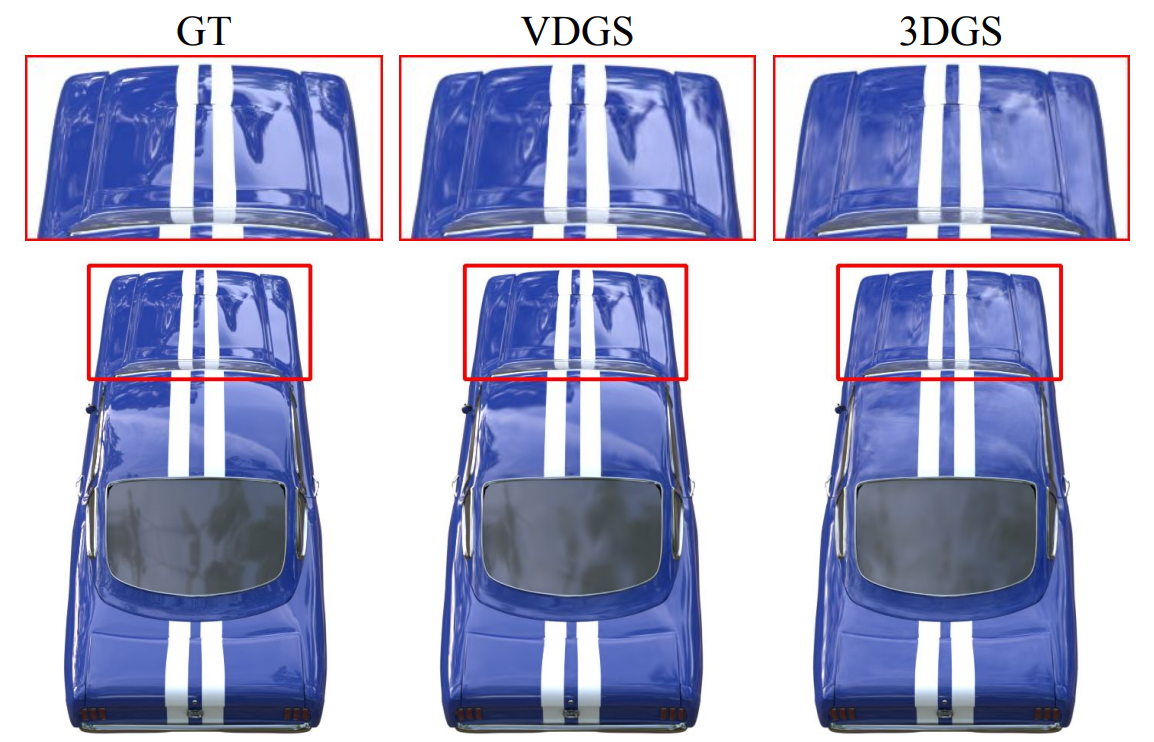}
    \caption{Visual comparison between classical GS and \our{} on the car scene from Shiny Blender \cite{verbin2022ref}.}
    \label{fig:shiny2}
\end{figure}

That is why, in this paper, we present a combination of the NeRF model and GS representation of 3D objects, which results in utilising the advantages of both these methods.

\section{\our{}}
\label{sec:method}


Here, we briefly describe the NeRF and GS models to better ground our findings. Next, we provide the details about \our{}, which is a hybrid of the above models. 

\subsection{NeRF representation of 3D objects}

Vanilla NeRF~\cite{mildenhall2020nerf} represents geometrically-rich 3D scenes using neural model taking as an input a 5D coordinate composed of the viewing direction ${\bf d} = (\theta, \psi)$ and spatial location $ \x = (x, y, z)$, that returns as its output emitted color ${\bf c} = (r, g, b)$ and volume density $\sigma$.

A vanilla NeRF employs a collection of images for training. This process entails generating an ensemble of rays that pass through the image and interact with a 3D object represented by a neural network. These interactions, including the observed color and depth, are then used to train the neural network to accurately represent the object's shape and appearance. NeRF approximates this 3D object with an MLP network: $$ \F_{NeRF} (\x , {\bf d}; \Theta ) = ( {\bf c} , \sigma). $$ The MLP parameters, $\bf \Theta$, are optimized during training to minimize the discrepancy between the rendered images and the reference images from a given dataset. Consequently, the MLP network takes a 3D coordinate as input and outputs a density value along with the radiance (i.e., color) in a specified direction.





Such approaches to representing 3D objects using neural networks often face limitations due to their capacity or difficulty in precisely locating where camera rays intersect with the scene geometry. Consequently, generating high-resolution images from these representations often necessitates computationally expensive optical ray marching, hindering real-time use case scenarios.

\subsection{Gaussian Splatting}

Gaussian Splatting (GS) model 3D scene by a collection of 3D Gaussians defined by a position (mean), covariance matrix, opacity, and color represented via spherical harmonics (SH) \cite{fridovich2022plenoxels,muller2022instant}.

The GS algorithm creates the radiance field representation by a sequence of optimization steps of 3D Gaussian parameters (i.e., position, covariance, opacity, and SH colors). The key to the GS's efficiency is the rendering process, which uses projections of Gaussian components. 

In GS, we use a dense set of 3D Gaussians: 
$$
\G = \{ (\N(\m_i,\Sigma_i), \sigma_i, c_i) \}_{i=1}^{n},
$$
where $\m_i$ is position, $\Sigma_i$ is covariance, $\sigma_i$ is opacity, and $c_i$ is SH colors of $i$-th component.

GS optimization is based on a cycle of rendering and comparing the generated image to the training views in the collected data. Unfortunately, 3D to 2D projection can lead to incorrect geometry placement. Therefore, GS optimization must be able to create, destroy, and move geometry if it is incorrectly positioned. The quality of the parameters of the 3D Gaussian covariances is essential for the representation's compactness, as large homogeneous areas can be captured with a small number of large anisotropic Gaussians.



\begin{figure*}
    \centering

    \includegraphics[width=0.8\textwidth, trim=0 0 0 0, clip]{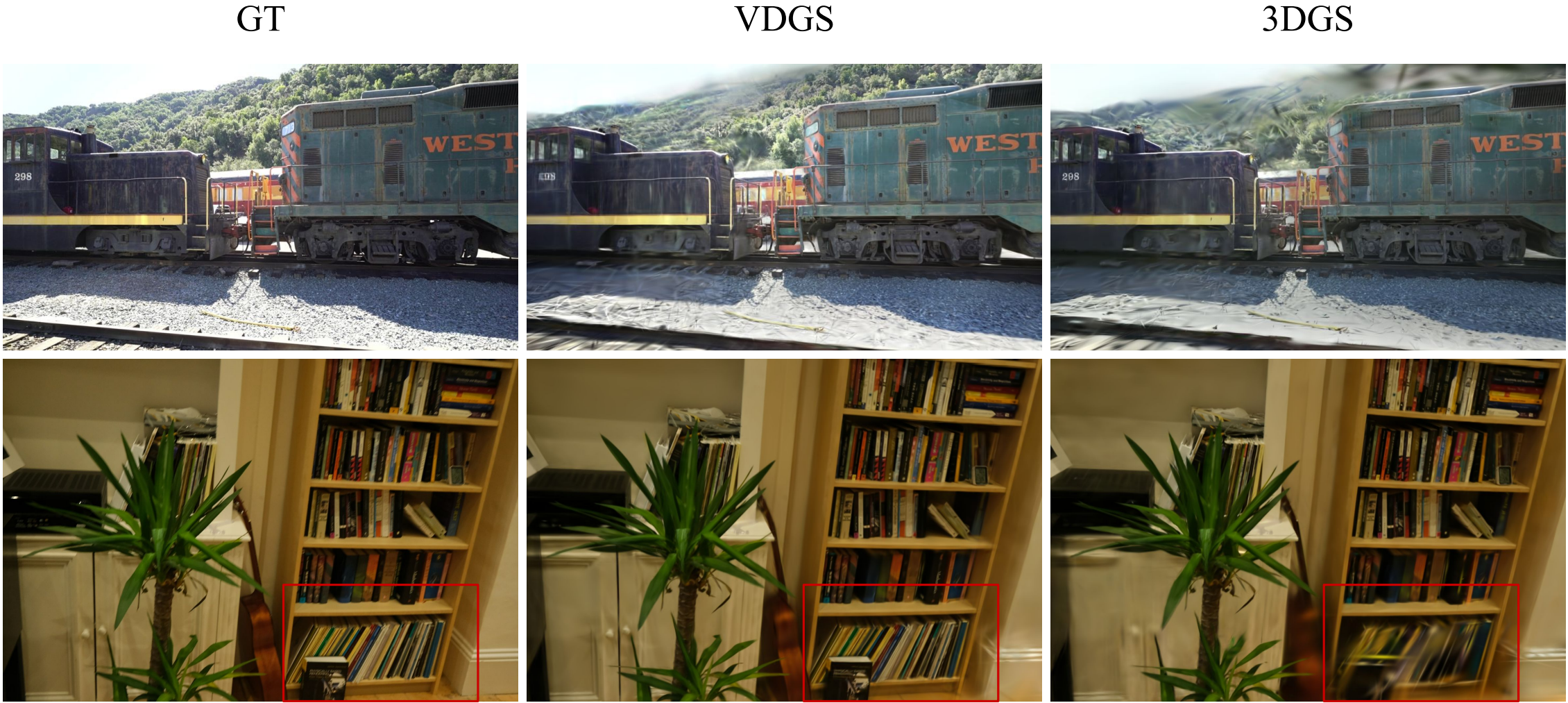}
    
    \caption{As our model is more susceptible to viewing direction, \ourshort{} can showcase the background more precisely. The present renders were made using Mip-NeRF 360 \cite{barron2022mip} and Tanks and Temples \cite{knapitsch2017tanks}.}
    \label{fig:background}
\end{figure*}


\subsection{\our{}}

In our approach, we use GS to model 3D shapes, while a NeRF-based neural network produces updates to colors and opacity based on viewing direction. Our model consists of the following GS representation:
$
\G
$
and the MLP network:
$$
\F_{\ourshort{}} (\m, {\bf d}; \Theta ) =   \Delta \sigma ({\bf d}),
$$
which takes the mean of Gaussian distribution and viewing direction ${\bf d}$ to returns 
volume density update $  \Delta \sigma ({\bf d}) $.
To better represent Gaussian's position, we use hash encoding \cite{muller2022instant} on means of Gaussians before taking them as input to our MLP. As it was shown in original NeRF paper \cite{mildenhall2020nerf} a crucial part for the best performance of NeRF-based models is taking coordinates into higher dimensional space using e.g. a frequency encoding. However, hash encoding proved to perform superiorly in this task. Not only does it give better results, but it also has extremely fast queries. That is why we have chosen this encoding type on Gaussian means.
The model is parameterized by $\bf \Theta$ and trained to change opacity depending on viewing direction. The final \ourshort{} model is given by:
$$
(\G, \F_{\ourshort{}} )= \{ (\N(\m_i,\Sigma_i), \sigma_i \cdot \Delta \sigma_i ({\bf d}) , {\bf c_i}  ) \}_{i=1}^{n},
$$
where 
$
\Delta \sigma_i ({\bf d})  = \F_{\ourshort{}} (\m, {\bf d}; \Theta ).
$

Consequently, we have two components, i.e., GS and NeRF. The former is used to produce the 3D object's shape and color. All Gaussian components have transparency values that do not depend on the viewing direction. This is why we also added the NeRF component, which can alter transparency values depending on the camera position. As a consequence, our model can add viewing direction-based changes. Therefore, it allows for the modeling of light reflection and transparency of objects. Moreover, it helps eliminate view-dependent artifacts and more precisely regulate background (see Figure \ref{fig:background}).

\begin{table*}[]
{
\begin{center}
    \begin{tabular}{l|lll|lll|lll}
    &  \multicolumn{3}{c}{Mip-NeRF360} & \multicolumn{3}{c}{Tanks\&Temples} & \multicolumn{3}{c}{Deep Blending} \\
         & SSIM $\uparrow$& PSNR $\uparrow$& LPIPS $\downarrow$& SSIM $\uparrow$& PSNR $\uparrow$& LPIPS $\downarrow$ & SSIM $\uparrow$& PSNR $\uparrow$& LPIPS $\downarrow$
 \\ \hline

Plenoxels & 0.626 & 23.08 &  0.719 & 0.379& 
21.08& 0.795 & 
0.510& 23.06& 0.510
\\
INGP-Base & 0.671 & 25.30 & 0.371 & 0.723 & 21.72 & 0.330 & 0.797 & 23.62 & 0.423
\\
INGP-Big & 0.699 & 25.59 & 0.331 & 0.745 & 21.92 & 0.305 & 0.817 & 24.96 & 0.390
\\
M-NeRF360 & \yellowc0.792 & \redc 27.69 & \yellowc0.237 & 0.759 & \yellowc22.22 & \yellowc0.257 & \yellowc0.901 & \yellowc29.40 & \orangec0.245
\\
\hline
Gaussian Splatting-7K & 0.770 & 25.60 & 0.279 & \yellowc0.767 & 21.20 & 0.280 & 0.875 & 27.78 & \yellowc0.317
\\
Gaussian Splatting-30K & \redc 0.815 & \yellowc 27.21 & \redc 0.214  & \orangec0.841 & \orangec 23.14 & \orangec0.183 & \orangec 0.903 & \orangec29.41 & \redc 0.243\\

\ourshort{} (Our)& \orangec 0.813& \orangec27.64& \orangec0.220& \redc 0.851 & \redc 24.02 & \redc 0.176 & \redc 0.906 & \redc 29.54 & \redc 0.243

\\ \hline
     
    \end{tabular}
    \end{center}
    }
    
    \caption{Quantitative evaluation of \our{} method compared to previous work, computed over three distinct datasets, namely, Mip-NeRF 360 \cite{barron2022mip}, Tanks and Temples \cite{knapitsch2017tanks}, and Deep Blending.}
    \label{tab:scene_mip_tt_db}
\end{table*}

Figure~\ref{fig:schemma} illustrates the architecture of \ourshort{}. The optimization starts with Structure from Motion (SfM) points, either obtained from COLMAP or generated randomly, setting the initial conditions for the 3D Gaussians. The camera position, as well as the hashed center of the Gaussian, are input into an MLP network to modify the opacity $\delta \sigma$ of the 3D Gaussians. Then, a differential Gaussian rasterization process is used to simultaneously optimize the MLP and 3D Gaussian parameters.

\ourshort{} computes the color $C(p)$ of a pixel by blending $N$ ordered points that overlap it, which are sampled from respective Gaussian distributions $\N(\m_i,\Sigma_i)$ for $i\in \{1,\ldots,N\}$. Precisely, the exact value of $C$ equals:
$$
C = \sum_{i \in N} c_i \alpha_i \prod_{j=1}^{i-1} (1-\alpha_i)
$$
where 
$$
\alpha_i = (1 - \exp(  \sigma_i \cdot  \F_{\ourshort{}} (\m_i, {\bf d}; \Theta ) \cdot  \delta_i)),
$$
$\delta_i$ is the distance between adjacent samples, and $\sigma_i$ denotes the opacity of sample $i$ multiply by NeRF-based function 
$\F_{\ourshort{}} (\m_i, {\bf d}; \Theta ),$ 
which depends on viewing direction $\d$.

All components are jointly trained, i.e., position (mean of Gaussian component), covariance matrix ($3 \times 3$ matrix), color (RGB) and opacity (transparency) multiplied by a NeRF-based function that depends on the viewing direction. 
GS starts with a limited number of points and then employs a strategy that involves creating new and eliminating unnecessary components. In practice, GS eliminates any Gaussians with an opacity lower than a specific limit after every hundred iterations. At the same time, new Gaussians are generated in unoccupied areas of 3D space to fill vacant areas. This GS optimization process is somewhat complex, but it can be trained very effectively thanks to robust implementation and the usage of CUDA kernels.

Our NeRF-based network can easily work in a GS setting by working with existing Gaussian components. Our network is tiny as we use only two layers with $32$ neurons. Consequently, \ourshort{} works extremely fast. Moreover, \ourshort{} has comparable training and inference time to vanilla GS.

\section{Experiments}
\label{submission}

In this section we show numerical experiments. To comprehensively validate the effectiveness of \ourshort{}, we conducted an evaluation on various datasets: a)
widely-used NVS dataset, i.e., the \textit{NeRF Synthetic} \citep{mildenhall2020nerf}, b) real-world, large-scale scenes dataset, i.e., the \textit{Tanks and Temples} \citep{knapitsch2017tanks}, \textit{Mip-NeRF 360} \citep{barron2022mip}, \textit{Deep Blending}  and
c) reflective objects datasets, i.e., the \textit{Shiny Blender} \citep{verbin2022ref}.

\paragraph{Quantitative results.} In almost all of the realistic scenes, \ourshort{} obtained the best score or a result slightly worse than Mip-Nerf 360 \citep{barron2022mip} (see Table \ref{tab:scene_mip_tt_db}). We also got almost all the highest PSNR values on single object datasets such as Synthetic (see Table \ref{tab:nerf}). In addition, we also achieved better scores than the original GS on the \textit{Shiny Blender} \citep{verbin2022ref} dataset (see Table \ref{tab:shiny}). In our approach, we do not model shadows and reflections directly. Thus, we do not obtain as good results as GaussianShader \cite{jiang2023gaussianshader}, which mainly focuses on properly representing shiny surfaces. However, our model is more flexible and helps with a wider range of challenges, which will be discussed in more detail in the subsequent paragraph.

\begin{figure*}
    \centering
    \includegraphics[width=0.9\textwidth, clip]{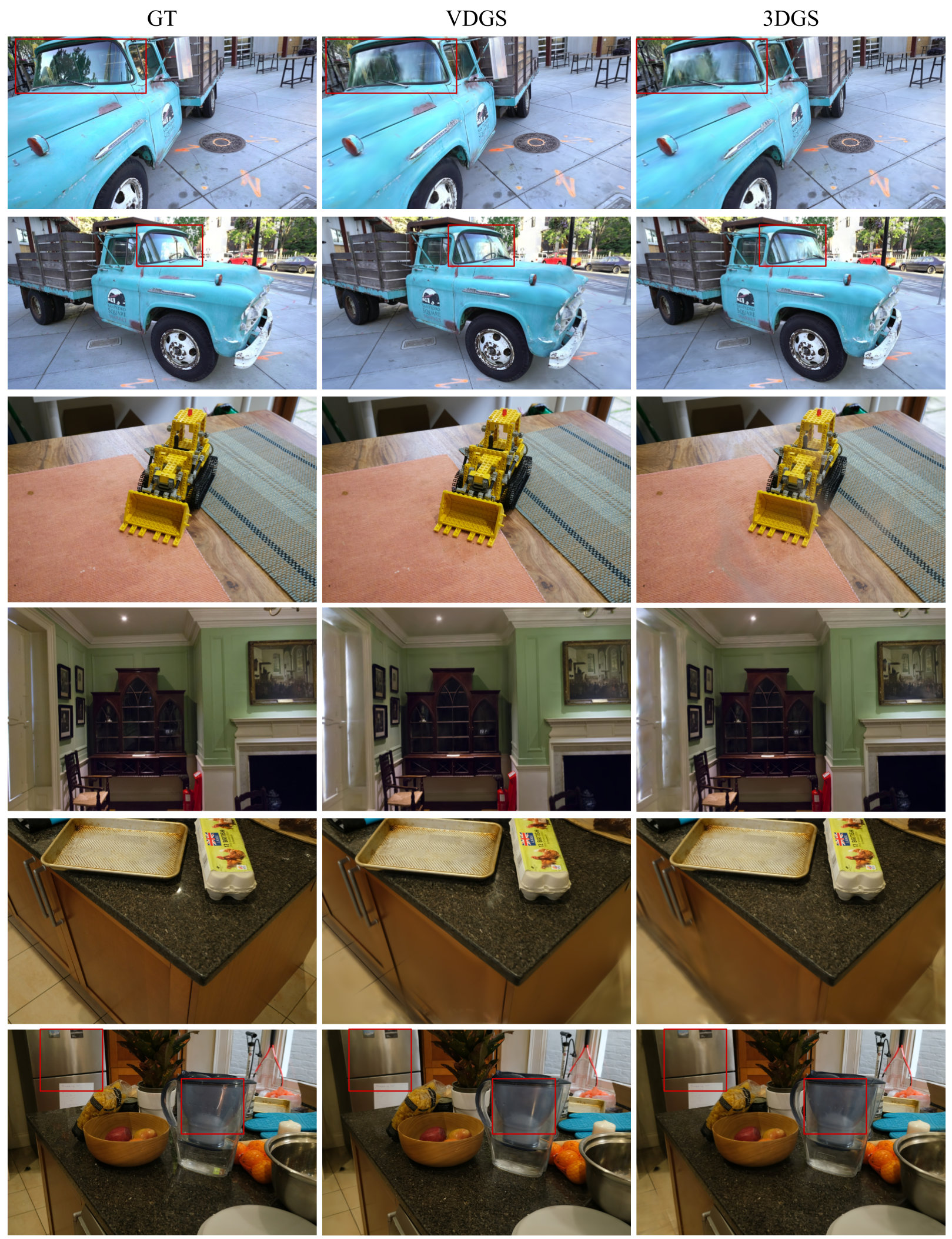}
    \caption{Further examples of how \ourshort{} can better model shiny surfaces and eliminate artifacts on realistic scenes from Mip-NeRF 360 \cite{barron2022mip}, Tanks and Temples \cite{knapitsch2017tanks} and Deep Blending.}
     \label{fig:extra1}
     \vspace{-0.3cm}
\end{figure*}

\begin{figure*}
    \centering
    \includegraphics[width=0.9\textwidth, clip]{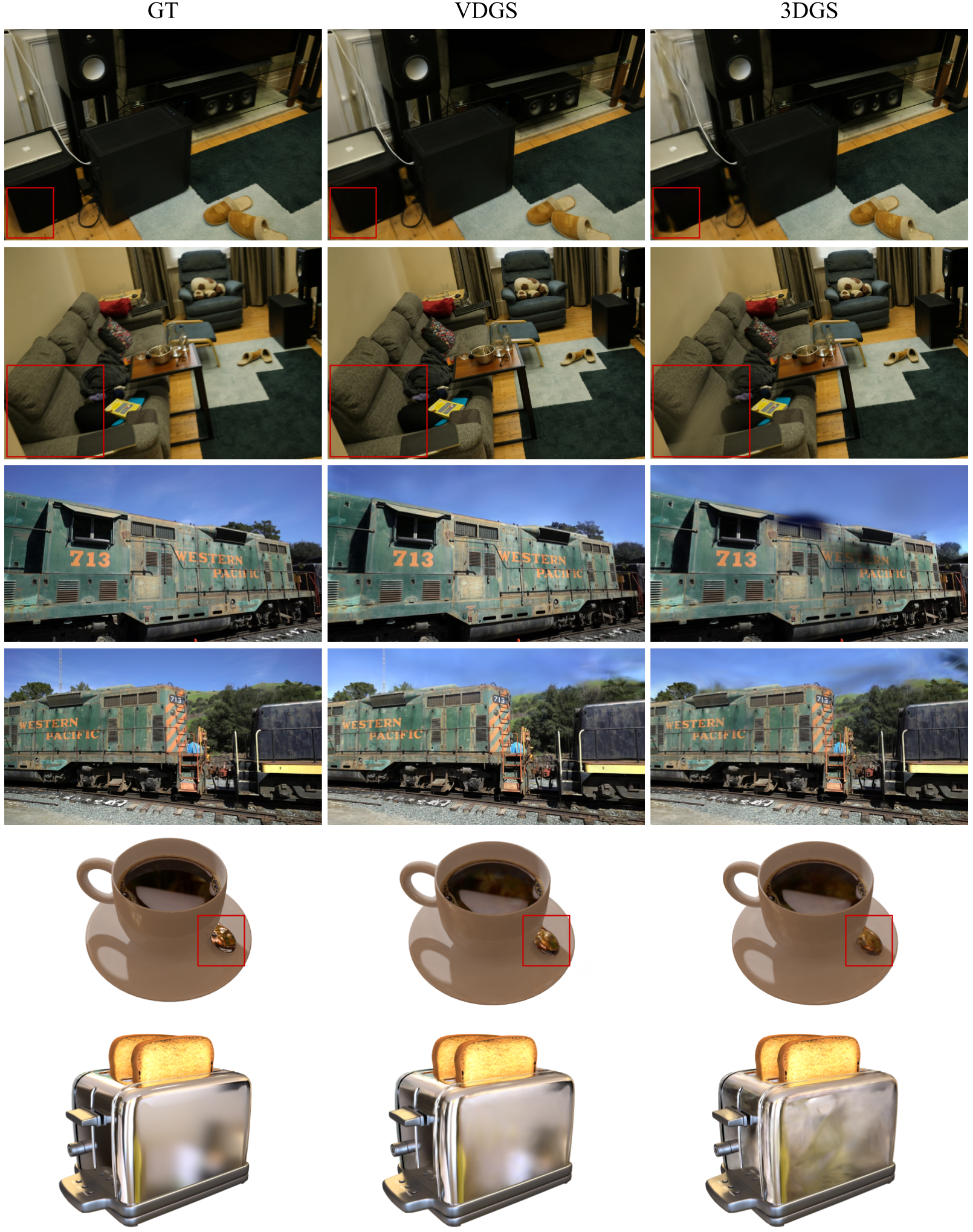}
    \caption{Visual comparison showcasing \ourshort{} ability to remove artifacts and more accurately represent background on realistic scenes from Mip-NeRF 360 \cite{barron2022mip} and Tanks and Temples \cite{knapitsch2017tanks}. Also, two examples of more precise representations of shiny surfaces on the Shiny Blender dataset \cite{verbin2022ref}.}
     \label{fig:extra2}
     \vspace{-0.3cm}
\end{figure*}

\paragraph{Quality results.} \our{} can accurately model glass surfaces and shiny objects both on real scenes (see Figure \ref{fig:merf_appro}, Figure \ref{fig:shiny1}, Figure \ref{fig:shiny_real}) and synthetic data (see Figure \ref{fig:shiny2}, Figure \ref{fig:shiny3}). Our model is also capable of removing artifacts that can show up in a specific viewing direction (Figure \ref{fig:artefacts_1}) as well as more accurately representing the background (see Figure \ref{fig:background}). More Figures illustrating the plausible quality of our results are available on the last two pages of this paper (see Figure~\ref{fig:extra1} and Figure~\ref{fig:extra2}).


\begin{table}[]
{
\scriptsize
\begin{center}
    \begin{tabular}{@{}l|lllllllll@{}}
    \multicolumn{9}{c}{PSNR $\uparrow$} \\
         & Chair & Drums & Lego & Mic & Mat. & Ship & Hot. & Ficus & Avg. \\ \hline
    NeRF & 33.00 & \yellowc25.01 & 32.54 & 32.91 & \yellowc 29.62 & 28.65 & 36.18 & \yellowc 30.13 & 31.01 \\
    VolSDF & 30.57 & 20.43 & 29.46 & 30.53 & 29.13 & 25.51 & 35.11 & 22.91 & 27.96 \\ 
    Ref-NeRF & \yellowc 33.98 & \orangec 25.43 & \yellowc 35.10 & \yellowc 33.65 & 27.10 & \yellowc 29.24 & \yellowc 37.04 & 28.74 & \yellowc 31.29 \\ 
    ENVIDR & 31.22 & 22.99 & 29.55 & 32.17 & 29.52 & 21.57 & 31.44 & 26.60 & 28.13 \\ \hline
    GS & \orangec 35.82 & \redc 26.17 & \redc 35.69 & \redc 35.34 & \orangec 30.00 & \orangec 30.87 & \orangec 37.67 & \orangec 34.83 & \orangec 33.30 \\ 
    \ourshort{}  (Our) & \redc 35.97 & \redc 26.17 & \orangec 35.40 & \orangec 34.76 & \redc 30.67 & \redc 30.94 & \redc 38.04 & \redc 34.98 & \redc 33.37 \\ \hline

    \multicolumn{9}{c}{SSIM $\uparrow$} \\
NeRF & 0.967 & 0.925 & 0.961 & 0.980 & 0.949 & 0.856 & \yellowc0.974 & \orangec0.964 & \yellowc0.947 \\
VolSDF  & 0.949 & 0.893 & 0.951 & 0.969 & 0.954 & 0.842 & 0.972 & 0.929 & 0.932 \\
Ref-NeRF & \yellowc 0.974 & 0.929 & \yellowc 0.975 & 0.983 & 0.921 & \yellowc 0.864 & \orangec 0.979 & \yellowc 0.954 & \yellowc 0.947 \\
ENVIDR & \orangec 0.976 & \yellowc 0.930 & 0.961 & \yellowc 0.984 & \redc 0.968 & 0.855 & 0.963 & \redc 0.987 & \orangec 0.956 \\ \hline
GS & \redc  0.987 & \redc 0.954 & \redc 0.983 & \redc 0.991 & \yellowc 0.960 & \redc 0.907 & \redc 0.985 & \redc 0.987 & \redc 0.969 \\
\ourshort{}  (Our) & \redc 0.987 & \orangec 0.950 & \orangec 0.981 & \orangec 0.990 & \orangec 0.965 & \orangec 0.903 & \redc 0.985 & \redc 0.987 & \redc 0.969
\\ \hline
    \multicolumn{9}{c}{LPIPS $\downarrow$ }\\ 
NeRF & 0.046 & 0.091 & 0.050 & 0.028 & 0.063 & 0.206 & 0.121 & \yellowc0.044 & 0.081 \\
VolSDF & 0.056 & 0.119 & 0.054 & 0.191 & 0.048 & 0.191 & 0.043 & 0.068 & 0.096 \\
Ref-NeRF & \yellowc 0.029 & \yellowc 0.073 & \yellowc 0.025 & \yellowc 0.018 & 0.078 & \yellowc 0.158 & \yellowc 0.028 & 0.056 & \yellowc 0.058 \\
ENVIDR & 0.031 & 0.080 & 0.054 & 0.021 & \yellowc 0.045 & 0.228 & 0.072 & \redc 0.010 & 0.067 \\ \hline
GS & \redc 0.012 & \redc 0.037 & \redc 0.016 & \redc 0.006 & \orangec 0.034 & \redc 0.106 & \redc 0.020 & \orangec 0.012 & \redc 0.030 \\
\ourshort{} (Our) & \orangec 0.013 & \orangec 0.042 & \orangec 0.018 & \orangec 0.008 & \redc 0.032 & \orangec 0.113 & \orangec 0.022 & \orangec 0.012 & \orangec 0.032

\\ \hline
    \end{tabular}
    \end{center}
    }
    
    \caption{The quantitative comparisons (PSNR / SSIM / LPIPS) on NeRF Synthetic dataset. }
    \label{tab:nerf}
\end{table}

\begin{table}[]
{
\scriptsize
\begin{center}
    \begin{tabular}
    {@{}l|lllllll@{}}
    \multicolumn{7}{c}{PSNR $\uparrow$} \\
         & Car & Ball & Helmet & Teapot & Toaster & Coffee & Avg.
 \\ \hline

NVDiffRec & \yellowc27.98 & 21.77 & 26.97 & 40.44 & 24.31 & 30.74 & 28.70
 \\
NVDiffMC & 25.93 & \yellowc30.85 & 26.27 & 38.44 & 22.18 & 29.60 & 28.88
 \\
Ref-NeRF & \redc 30.41 & 29.14 & \orangec29.92 & 45.19 & 25.29 & \redc 33.99 & \orangec32.32
 \\
NeRO & 25.53 & 30.26 & \yellowc29.20 & 38.70 & \redc 26.46 & 28.89 & 29.84
 \\
ENVIDR  &  \orangec 28.46 & \redc 38.89 & \redc  32.73 & 41.59 & \yellowc26.11 & 29.48 & \redc 32.88
 \\
 \hline
GS & 27.24 & 27.67 & 28.28 & \yellowc45.70 & 21.01 & \yellowc32.30 & 30.37  \\
GaussianShader & 27.90 & \orangec 30.98 & 28.32 & \redc 45.86 &   \orangec26.21 & \orangec32.39 & \yellowc31.94 \\
\ourshort{} (Our) & 26.86 & 28.54 & 28.56 & \orangec45.84 & 21.71 & 32.19 & 30.62 \\ \hline

    \multicolumn{7}{c}{SSIM $\uparrow$} \\
NVDiffRec & \redc 0.963 & 0.858 & 0.951 & \orangec 0.996 & \yellowc0.928 & \redc 0.973 & 0.945 \\
NVDiffMC & 0.940 & 0.940 & 0.940 & \yellowc0.995 & 0.886 & 0.965 & 0.944 \\
Ref-NeRF & \yellowc 0.949 & 0.956 & \yellowc 0.955 & \yellowc0.995 & 0.910 & \orangec 0.972 & 0.956 \\
NeRO & \yellowc 0.949 & \orangec 0.974 & \orangec 0.971 & \yellowc0.995 & \orangec 0.929 & 0.956 & \orangec 0.962 \\
ENVIDR & \orangec 0.961 & \redc 0.991 & \redc 0.980 & \orangec 0.996 & \redc 0.939 & 0.949 & \redc 0.969 \\
\hline
GS & 0.931 & 0.937 & 0.951 & \redc0.997 & 0.895 & \orangec0.972 & 0.947 \\
GaussianShader  & 0.931 & \yellowc 0.965 & 0.950 & \orangec 0.996 & \orangec 0.929 & \yellowc 0.971 & \yellowc 0.957 \\
\ourshort{}  (Our) & 0.925 & 0.944 & 0.950 & \redc 0.997 & 0.896 & \yellowc 0.971 & 0.947

\\ \hline
    \multicolumn{7}{c}{LPIPS $\downarrow$ }\\ 
NVDiffRec & \redc 0.045 & 0.297 & 0.118 & \orangec0.011 & 0.169 & \redc 0.076 & 0.119 \\
NVDiffMC & 0.077 & 0.312 & 0.157 & 0.014 & 0.225 & 0.097 & 0.147 \\
Ref-NeRF & 0.051 & 0.307 & 0.087 & 0.013 & 0.118 & \yellowc0.082 & 0.109 \\
NeRO & 0.074 & \orangec 0.094 & \redc 0.050 & \yellowc0.012 & \orangec 0.089 & 0.110 & \orangec 0.072 \\
ENVIDR & \yellowc0.049 & \redc 0.067 & \orangec 0.051 & \orangec0.011 & 0.116 & 0.139 & \orangec 0.072 \\
\hline
GS & \orangec0.048 & 0.162 & 0.080 & \redc0.007 & 0.126 & \orangec0.078 & 0.083 \\
GaussianShader & \redc 0.045 & \yellowc 0.121 & 0.076 & \redc 0.007 & \redc 0.079 & \orangec 0.078 & \redc 0.068 \\

\ourshort{} (Our) & 0.050 & 0.142 & \yellowc 0.072 & \redc 0.007 & \yellowc 0.107 & \orangec 0.078 & \yellowc0.076

\\ \hline
    \end{tabular}
    \end{center}
    }
    
    \caption{Quantitative comparisons (PSNR / SSIM / LPIPS) in the Shiny Blender dataset \cite{verbin2022ref} with NVDiffRec \cite{munkberg2022extracting}
NVDiffMC \cite{hasselgren2022shape}
Ref-NeRF \cite{verbin2022ref}
NeRO \cite{liu2023nero}
ENVIDR \cite{liang2023envidr}
Guassian Splatting  \cite{kerbl20233d}
GaussianShader \cite{jiang2023gaussianshader}.
 }
    \label{tab:shiny}
\end{table}

\begin{table}
{
\scriptsize
\begin{center}
    \begin{tabular}{l|ccccccc}
         & Car & Ball & Helmet & Teapot & Toaster & Coffee & Avg. \\ \hline

GS & \orangec27.24 & \yellowc27.67 & \yellowc28.28 & \yellowc45.70 & \yellowc21.01 & \orangec32.30 & \yellowc30.37\\
\ourshort{} (Our) & \yellowc26.86 & \redc28.54 & \redc28.56 & \redc45.84 & \redc21.71 & \yellowc32.19 & \redc30.62 \\
Pre-\ourshort{} (Our) & \redc27.29 & \orangec27.81 & \orangec28.37 & \orangec45.71 & \orangec21.17 & \redc32.33 & \orangec30.45\\
\hline
    \end{tabular}
    \end{center}
     } 

     \caption{Comparison between original GS model 30k epochs, \ourshort{} 30k epochs and pre-trained GS standard model with additional 20k epochs with \ourshort{} on Shiny Blender \cite{verbin2022ref} data. We called the pre-trained model as Pre-\ourshort{}.}
     \label{tab:pretrained_synt}
\end{table}



\begin{table}[]
\scriptsize
\begin{center}
{
    \begin{tabular}{@{}l|ccccccccccc@{}} 
     \multicolumn{7}{c}{NeRF Synthetic PSNR $\uparrow$} \\
\ourshort{}  & Chair & Drums & Lego & Mic & Materials & Ship & Hotdog & Ficus & Avg. \\ \hline

 C$+$ &  35.87 & 26.31 & \orangec35.71 & \yellowc35.27 & \yellowc30.02 & 27.66 & \yellowc37.77 & \orangec35.08 & 32.96 \\

C$*$ & \yellowc35.91 & \orangec26.20 & \redc35.73 & \redc35.44 & 29.96 & \yellowc29.63 & 37.76 & \yellowc35.03 & \yellowc33.21 \\

O$+$ &  35.60 & 26.14 & 34.41 & \orangec35.33 & 29.97 & 28.70 & 37.09 & 34.93 & 32.77 \\	

O$*$  &  \orangec35.97 & \yellowc26.17 & 35.40 & 34.76 & \redc30.67 & \redc30.94 & \redc38.04 & 34.98 & \redc 33.37 \\

OC$+$ & 35.77 & \redc26.28 & 31.17 & 33.70 & 29.49 & 17.91 & 36.68 & \redc35.24 & 30.78 \\	
OC$*$  &  \redc36.21 & 26.03 & \yellowc35.44 & 34.59 & \orangec30.66 & \orangec30.88 & \orangec38.03 & 34.29 & \orangec33.27
    \end{tabular}
}

\end{center}
    \caption{Ablation study of different versions of \ourshort{}. We compare different models: Color Add (C$+$), Color Multiply C$*$, Opacity Add (O$+$), Opacity Multiply (O$*$) Opacity and Color Add (OC+), Opacity and Color Multiply (OC*). As we can see, different versions considered have good scores in various data sets and measures. In our paper, we chose VDGS Opacity Multiply (Our) as a final model since we obtained the largest number of best scores on the PSNR measure.}
    \label{tab:all_our}
\end{table}





\paragraph{Pre-trained GS}
Normally, our model simultaneously trains GS and NeRF components. In Table~\ref{tab:pretrained_synt}, we present results on the pre-trained classical GS model. We use a fully trained GS model, and then we additionally train our neural network. As can be seen, such an approach gives better results than the original GS. Furthermore, we obtain the best scores when we train both components simultaneously.

\paragraph{Implementation Details}
We implement \ourshort{} based on the original GS repository using the PyTorch and TinyCUDA frameworks. We trained our model on a single NVIDIA RTX 3090 GPU. We also utilized a 3D structured hash grid with 12 levels of 3D embedding ranging from 16 to 512 resolutions. The maximum hash table size was $2^{14}$ with a feature dimension of 2. For our NeRF-based network, we used a 2-layer MLP with LeakyReLU activation. We made sure to replicate the same configuration for all scenes.

\paragraph{Training and inference time}
In contrast to GS in \ourshort{}, we use a viewing direction neural network. Therefore, we have slightly longer training and inference time, as shown in Table~\ref{tab:my_time}. However, in practice, our model can work in real-time during inference. 

\paragraph{Ablation Study}
Our \ourshort{} updates only opacity. It is a natural question of how our model works when also color changes. In the ablation study, we verified different versions of \our{}. In Table~\ref{tab:all_our}, we list the PSNR values for a different version of our model. At the same time, in the appendix, we present the full results of PSNR/SSIM/LPIPS measures.

As we can see, different versions of \our{} result in high scores for various data sets. In our paper, we chose VDGS Opacity Multiply as a final model since we obtained the highest number of best scores on the PSNR measure on the data containing real scenes. It should also be highlighted that the model VDGS Color Multiply gets the best scores on the NeRF Synthetic dataset. Models that use both color and opacity also obtain good results but are slightly worse than models that use only one of such elements. We believe this effect may be attributed to the dependency of opacity on color. Therefore, joint training is not effective.


\begin{table}
{\scriptsize
    \begin{center}
    \begin{tabular}{@{}c@{\;}|@{\;}c@{\;}c@{\;}c@{\;}c@{\;}c@{\;}c@{}} 
        &  \multicolumn{2}{c}{Mip-NeRF360} & \multicolumn{2}{c}{Tanks\&Temples} & \multicolumn{2}{c}{Deep Blending}\\ 
         & Train & FPS & Train & FPS & Train & FPS  \\ \hline 
         Plenoxels & 25m49s & 6.79 & 25m5s & 13.0 & 27m49s & 11.2 \\
         INGP-Base & 5m37s & 11.7 & 5m26s & 17.1 & 6m31s & 3.26 \\
         INGP-Big & 7m30s & 9.43 & 6m59s & 14.4 & 8m & 2.79 \\
         M-NeRF360 & 48h & 0.06 & 48h & 0.14 & 48h & 0.09 \\ 
         GS-30K &  41m33s & 134 & 26m54s & 154 &  36m2s & 137 \\ \hline
         GS-30K* &  26m21s & 89.28 & 14m50s & 94.07 & 24m04s  & 87.95 \\
         \ourshort{}-30K* (Our) &  46m11s & 41.35 & 29m45s & 38.53 & 41m29s & 44.72 \\

    \end{tabular}
    \end{center}
    \caption{Comparison of training time and frame rate. We train \ourshort{} on the RTX 3090 GPU. We reproduce the result of GS on the RTX 3090 GPU. \ourshort{} has slightly slower training and inference time compared to GS. However, it still yields real-time FPS superior to previous NeRF-based methods.}
    \label{tab:my_time}
}
\end{table}

\section{Limitations}

Our model can outperform the original GS at multiple tasks, such as rendering glass surfaces and shadows or eliminating artifacts. However, due to its generality, it cannot reproduce the results of other solutions specifically designed to tackle selected niche problems, like accurate light reflections, by better simulating physical attributes.

\section{Discussion and Conclusions}
This paper presents a new neural rendering strategy that leverages two main concepts, NeRF and GS. We represent a 3D scene with a set of Gaussian components and a neural network that can change the color and opacity of Gaussians concerning viewing direction. This approach inherited the best elements from the NeRF and GS methods. We observed rapid training and inference, and we can model shadows, light reflections, and transparency with a quality similar to that of NeRF-based models. Furthermore, the experiments conducted show that \our{} gives superior results than NeRF and the GS model.



\nocite{langley00}

%

\section{Appendix}

\end{document}